\documentclass[10pt]{article} 
\usepackage[preprint]{tmlr}
\usepackage{graphicx}

\usepackage{amsmath,amsfonts,bm}

\def\figref#1{figure~\ref{#1}}

\def\secref#1{section~\ref{#1}}

\def\eqref#1{equation~\ref{#1}}

\def\1{\bm{1}}

\DeclareMathAlphabet{\mathsfit}{\encodingdefault}{\sfdefault}{m}{sl}
\SetMathAlphabet{\mathsfit}{bold}{\encodingdefault}{\sfdefault}{bx}{n}

\usepackage{hyperref}
\usepackage{url}
\usepackage{array}
\usepackage{booktabs}
\usepackage{multirow}
\usepackage{enumitem}
\usepackage{listings}
\title{
The (Butter)fly Dream: Linear Separability of World-Model Rollouts in a Connectome-Constrained Visual Model
}

\author{
  \name Junichiro Niimi
  \email jniimi@meijo-u.ac.jp \\
  \addr Faculty of Business Management\\ Meijo University
}

\providecommand{\tabref}[1]{table~\ref{#1}}
\providecommand{\Tabref}[1]{Table~\ref{#1}}

\begin{document}

\maketitle

\begin{abstract}
Zhuangzi dreamt he was a butterfly and, on waking, could not tell which was the dream. We ask a version of his question of a model of the fly brain: placed inside a world generated by a world model, can a fixed brain tell that the world is a dream? We treat it as a question about what a connectome-constrained model trained on one task makes available beyond that task: do its responses make the defects of a dream linearly decodable? The observer is a published model of the fly visual system trained only for optic-flow estimation; reality is the MPI-Sintel movie on the model's 721-column retina; dreams are continuations of five real frames by three small world models (a persistence random walk, an MDN-RNN and an RSSM) at three temperatures. The judge is a classifier two-sample test on temporal statistics of cell-type responses, with sequence-level cross-validation. In the trained observer's responses all nine dreams are separable from reality (AUC 0.85--0.99, mean over 50 trained models), in those of the same network before training much less so (0.58--0.75 for the learned world models; paired difference 0.20--0.37), equally on sequences the observer never saw, and in an order unrelated to pixel error. Because these AUCs are near ceiling, we add graded probes made of each window's own real frames with only their order broken. Swapping one adjacent pair of the 14 generated frames gives 0.73 on the trained responses against 0.54 before training and 0.56 for pixel statistics; the trained curve reaches 0.75 at 1.1 swapped pairs of seven, the pixel statistics at 4.7, and the untrained network not at all. A detection horizon, the test on the first $k$ generated frames, orders the world models where the full-length AUC saturates.
\end{abstract}

\section{Introduction}\label{sec:intro}

\begin{quote}
Once Chuang Chou dreamt he was a butterfly, a butterfly flitting and fluttering around, happy with himself and doing as he pleased. He didn't know he was Chuang Chou. Suddenly he woke up and there he was, solid and unmistakable Chuang Chou. But he didn't know if he was Chuang Chou who had dreamt he was a butterfly, or a butterfly dreaming he was Chuang Chou. Between Chuang Chou and a butterfly there must be some distinction! This is called the Transformation of Things.

\hfill --- \textit{Zhuangzi}, trans.\ Burton Watson \citeyearpar[p.~49]{zhuangzi_watson}
\end{quote}

In this fable, the Chinese philosopher Zhuangzi (Chuang Chou in Watson's romanization) cannot tell the dream from waking. We put the question to a model of the fly brain, in a setting in which the dream can be built and measured: the brain is fixed, the world it sees is generated by a world model from a few frames of a real movie, and we ask whether the brain can tell.

The question concerns the brain model rather than the world model. Connectome-constrained models of the fly visual system take their wiring from the connectome, are trained on a single task and are validated against measured responses of their cell types \citep{flyvis}; what such training gives a model beyond its task is a question in its own right, and the released models are themselves under scrutiny \citep{flyhack,zhou2026geometry}. A world model supplies a natural test. Its continuations of real context resemble reality frame by frame but depart from it in their temporal structure, in ways that differ between world models and sampling temperatures, and the observer was never trained to notice such departures. This is not the role that world models play in the agents of \citet{ha2018} and the Dreamer line \citep{planet,dreamerv2}, where the world model lives inside the agent and its dreams serve control rather than fidelity: \citet{ha2018} raised the sampling temperature of their world model, making the dream less faithful to reality, so that the controller trained in it could not exploit the model's imperfections, and the controller then transferred better to the real environment. Whether a dream can be told from reality is therefore not a criterion of a world model's usefulness, and we do not use it as one. We use the dreams as stimuli for a fixed, published model of a biological visual system that was never trained on our data, never trained to detect anything, and cannot adapt to the generator's errors. If its responses to a dream are statistically indistinguishable from its responses to reality, the dream passes in that response space; if they differ in a way that a classifier trained on them can pick up, the observer's training has made the dream's departure from reality available in its responses, and the question becomes how quickly, and from how small a departure.

We make the question operational. The observer is the connectome-constrained network of the fly visual system of \citet{flyvis}, trained by its authors for optic-flow estimation and distributed as an ensemble of 50 independently trained models; we use it as a black box. Reality is luminance from the MPI-Sintel movie \citep{sintel}, rendered to the model's hexagonal retina with the model's own preprocessing. Dreams are continuations of five real frames by world models trained in the same 721-column space, so no renderer and no decoder stand between the generator and the observer. The judge is a classifier two-sample test \citep{c2st}: a logistic regression trained to separate the observer's responses to reality from its responses to a dream, using temporal statistics of its cell-type responses, evaluated with a cross-validation that keeps movie sequences disjoint between every stage. The observer itself decides nothing. What we measure is whether its response space makes the deviations of a dream from reality linearly decodable, and this is the precise form in which we answer whether the brain can tell.

Our contributions are the following.
\begin{enumerate}[leftmargin=*,itemsep=1pt]
\item A measurement protocol in which a fixed, published, connectome-constrained observer provides the response space in which world-model dreams conditioned on real context are compared with reality, with a two-sample test on temporal statistics of cell-type responses, sequence-level cross-validation and cluster bootstrap confidence intervals, and with two controls that fix what the observer adds: the same network at five initialisations, and the pixel statistics of the dream itself (\secref{sec:setup}).
\item The finding that all nine dreams (three world models at three temperatures) are linearly separable from reality in the trained observer's responses (AUC 0.85--0.99 averaged over the 50 models), that in the untrained network's responses the learned world models largely are not (0.58--0.75 over five seeds; paired cluster-bootstrap difference 0.20--0.37), that the separation is not confined to the sequences the observer was trained on, and that the ranking of the dreams is unrelated to their pixel error (Spearman $-0.07$ over the nine-condition grid, against $-0.68$ for the pixel-statistics control) (\secref{sec:results}).
\item Graded temporal probes that answer what the saturated dreams cannot, namely how much temporal disorder the trained responses need: continuations made of each window's own real frames, with $k$ of seven adjacent pairs swapped or $r$ of 14 frames repeated as slow motion, whose strength zero is reality itself. One swapped pair, which leaves the set of frames unchanged, gives an AUC of 0.73 on the trained responses against 0.54 for the untrained network and 0.56 for the pixel statistics; the trained curve reaches 0.75 at $k = 1.1$ [0.9, 1.4] against 4.7 [3.9, 5.4] for the pixel statistics, and the untrained network does not reach it at all; in slow motion the thresholds are $r = 2.7$, 6.0 and 8.2 (\secref{sec:probes}).
\item A detection horizon, the test restricted to the first $k$ generated frames, that separates the world models in time where the full-length AUC saturates: the persistence walk and the RSSM are separable from the first generated frame, while the MDN-RNN at low temperature accumulates separable evidence slowly and never reaches an AUC of 0.9 within the 14 generated frames (\secref{sec:horizon}).
\item A secondary observation: one-class density scores on the same features, a shrinkage Gaussian and a latent-space energy-based model, stay at AUC 0.56--0.69 and fall below 0.5 on smooth dreams; one-class typicality need not align with two-sample distinguishability (\secref{sec:density}).
\end{enumerate}

We do not claim that the connectome is what makes the observer's responses diagnostic. Our controls establish that training the connectome-constrained network makes the dreams' deviations linearly decodable from its responses and that this is not reducible to pixel statistics. Our question is what the responses of a model of a real brain, trained on one task, make available beyond that task, not which representation detects dreams best; we therefore compare the observer with itself before training and with the pixel statistics of the dream, not with generic learned detectors, whose separability would answer a different question (\secref{sec:discussion}).

\section{Related work}\label{sec:related}

\paragraph{Connectome-constrained models.} \citet{flyvis} built a network of the fly optic lobe whose cells, cell types and synaptic connectivity follow the connectome, with 734 free parameters (cell-type biases, time constants and synapse strengths) trained end to end on optic-flow estimation from naturalistic movies; the trained models reproduce measured responses of many cell types. We use their released ensemble of 50 models unchanged, their retinal rendering and their training data. The ensemble lets us report the spread of our measurement over independently trained brains. The released ensemble has itself become an object of study: \citet{flyhack} tests its mechanistic claims against causal temporal CNNs trained on the same optic-flow task, and \citet{zhou2026geometry} compare its representational geometry with sign-preserving shuffles of its parameters.

\paragraph{World models.} The MDN-RNN of \citet{ha2018}, a recurrent network with a mixture-density head \citep[MDN;][]{mdn} as used for sequence generation by \citet{graves2013generating}, and the recurrent state-space model~\citep[RSSM;][]{planet}, with the KL balancing and free bits of \citet{dreamerv2}, are the two learned generators we use. Surveys of world models are given by \citet{worldmodel_survey_mai} and \citet{worldmodel_survey_ding}. In those works the generator is an instrument for an agent; here it supplies the stimuli.

\paragraph{Evaluating generated video.} Generated video is commonly scored with distances between distributions of learned features, for instance the Fr\'echet video distance \citep{fvd}, whose feature extractor is a network trained for action recognition. Our observer is also a fixed feature extractor, but it was trained on a biological wiring diagram for a biological task, it has an explicit internal time (integration step 20 ms), and its features are responses of named cell types; our question is not whether a dream is close to real video in some feature space, but whether a particular biological system would respond to it as to reality.

\paragraph{Two-sample tests and the failure of likelihoods.} The classifier two-sample test \citep{c2st} measures the distinguishability of two samples by the held-out accuracy of a classifier trained to tell them apart. We use it as the primary measure; likelihood-based detectors, by contrast, can assign higher density to out-of-distribution inputs than to training data \citep{nalisnick2019}; we observe the same effect with smooth dreams, which look more typical than reality to a density fitted on reality.

\paragraph{Energy-based anomaly scores.} Energy-based models \citep{ebm_tutorial,du2019implicit} and in particular energies attached post hoc to a fixed encoder's latent space \citep{lebm} are a standard route to anomaly scores \citep{ebm_anomaly}. We include such a head, trained with short-run Langevin contrastive divergence \citep{du2019implicit}, as one of the density-based comparisons and report its scores as a secondary observation.

\section{Setup}\label{sec:setup}

\subsection{Observer: a fixed connectome-constrained visual system}\label{sec:observer}

The observer is the \texttt{flyvis} network of \citet{flyvis} (version 1.2.0), a hexagonal-lattice model of the fly optic lobe with 45,669 cells of 65 cell types, whose connectivity is fixed by the connectome and whose 734 free parameters were trained by its authors to predict optic flow from naturalistic movies. The input is luminance on 721 hexagonal columns (lattice extent 15), produced by the authors' \texttt{BoxEye} renderer: a $13 \times 13$ box filter sampled at the receptor centres. The network is integrated with a step of $\mathrm{d}t = 1/50$~s, the value used in training; a 24~fps movie is resampled to this step by the authors' nearest-exact interpolation, so 19 frames become 40 steps. Before every stimulus the state is initialised by the authors' fade-in procedure: the first frame is presented for 1~s with contrast ramped from grey. All simulations run on CPU, unmodified.

From the response of all cells we keep, per cell type, the mean over all cells of that type (\emph{type-mean}, 65 series) and the response of the cell in the central column (\emph{central}, 65 series); the main text uses the type-mean, the central variant is in the appendix. We use all 50 models of the released ensemble (\texttt{flow/0000/000}--\texttt{049}), which share the connectome and differ in initialisation; they are 50 independently trained brains.

\paragraph{Untrained control.} The released model directories contain only the final parameters (the single stored checkpoint is identical to the best checkpoint), so no pre-training snapshot exists. We therefore construct the untrained observer as the same network built from the model configuration without restoring any checkpoint: the cell-type biases are drawn from the configuration's initial distribution ($\mathcal{N}(0.5, 0.05^2)$ per type), the time constants take their initial value (0.05) and the synapse strengths their initial scaling (0.01 times the connectome's synapse counts). Signs and synapse counts, which are not trained, are the connectome's and identical to the trained models'. This is the connectome without training. The bias draw is the only random part of the initialisation, so we build five untrained observers with the configuration's seed set to 0--4; they differ only in the 65 cell-type biases, and the differences between them are correspondingly small.

\subsection{Reality: Sintel on the hexagonal retina}\label{sec:reality}

Reality is the MPI-Sintel movie \citep{sintel}, the data the observer was trained on, in its \emph{final} rendering pass, which is what \texttt{flyvis} uses. We use all 35 sequences for which frames are distributed: the 23 training sequences (1,064 frames) and the 12 test sequences (564 frames). \texttt{flyvis} fitted its parameters on 17 of the training sequences and kept the other six for validation; the 12 test sequences have no optical flow and were never used by \texttt{flyvis}. We call the 17 sequences \emph{observer-seen} and the other 18 \emph{observer-unseen}. Each sequence is converted to luminance, centre-cropped to 70\% of its width, cut into three overlapping vertical strips, and rendered to 721 columns with the authors' renderer and the authors' code path, so that our real stimuli match the observer's training preprocessing exactly (our rendering agrees with the authors' cached renderings to machine precision); the first frame of each sequence is skipped as in the authors' pipeline. This gives 105 strip series of 19--49 frames at 24~fps. Each series is tiled into non-overlapping windows of 19 frames (0.79~s), 198 windows in all, 96 of them from observer-seen and 102 from observer-unseen sequences. No augmentation and no noise are applied to reality. Sintel frames are not reproduced here; the dataset is used for academic purposes only.

\subsection{Dreams: three world models in the retina's own space}\label{sec:dreams}

A dream is a window in which the first $C = 5$ frames are real (the \emph{context}) and the remaining 14 frames are generated by a world model conditioned on the context, clipped to $[0,1]$. Because the brain's input closes in the 721-column space, the world models are trained directly in that space, with no renderer and no decoder in between. Three world models are compared.

\emph{Persistence random walk} (no learning, a lower bound): $x_t = \mathrm{clip}(x_{t-1} + \tau \sigma \epsilon_t)$ with $\epsilon_t \sim \mathcal{N}(0, I)$ per column, starting from the last context frame, where $\sigma$ is set once per cross-validation fold so that at $\tau = 1$ the expected frame-to-frame change $\sigma \sqrt{2/\pi}$ equals the real frame-to-frame change of the fold's training series ($\sigma$ = 0.048--0.062 across folds).

\emph{MDN-RNN} \citep{ha2018}: an LSTM with 256 hidden units and a mixture-density head \citep{mdn} with five diagonal Gaussian components over the 721 columns, with the current input fed to the head alongside the hidden state. It is trained on random 19-frame crops of the training series (batch 32, AdamW, learning rate $10^{-3}$, gradient clipping at 1, 60 epochs of 50 steps), with the authors' hexagonal rotations ($n \times 60^\circ$) and flips applied to training crops only, the geometric augmentation \texttt{flyvis} itself used; the checkpoint with the best validation negative log-likelihood is kept (best epochs 19--50 across folds). Generation samples a component and a Gaussian draw per frame; the temperature $\tau$ scales the mixture logits and the component standard deviations by $\sqrt{\tau}$ \citep{ha2018}.

\emph{RSSM} \citep{planet,dreamerv2}: an encoder (721--256--256, ELU), a GRU \citep{gru} deterministic state of 256 units, a 32-dimensional diagonal-Gaussian stochastic latent with prior from the state and posterior from the state and the encoded frame, and a decoder from state and latent to 721 columns with a Gaussian likelihood of fixed standard deviation 0.1, trained with KL balancing (0.8 on the prior side) and free bits (1~nat) as in \citet{dreamerv2}, with the same data pipeline and augmentation as the MDN-RNN and early stopping on the validation negative ELBO (best epochs 47--58). Generation runs the posterior through the context and then samples the prior with its standard deviation multiplied by $\tau$; the output is the decoder mean, so RSSM dreams carry no observation noise.

Each world model is run at $\tau \in \{0.25, 0.5, 1.0\}$, giving nine dreams per real window. For every dream we also record conventional error measures over the generated frames: the mean absolute pixel error to the real continuation, the persistence error (holding the last context frame), the mean frame-to-frame absolute change, and the spatial standard deviation per frame. The real windows have a frame-to-frame change of 0.041 and a persistence error of 0.108.

\subsection{Graded temporal probes}\label{sec:probes_setup}

The nine dreams depart from reality in many ways at once, and all of them turn out to be separable in the trained responses (\secref{sec:results}), so they do not show how large a departure the response space needs. We therefore also build continuations whose generated frames are the window's own real frames, so that every frame is a real frame and only the temporal order is broken, with a strength that can be set from zero. No world model is involved: the probe of a window replaces generated frame $t = 1, \dots, 14$ (real frame $4 + t$) by real frame $\mathrm{idx}(t)$ of the same window, without interpolation, and keeps the five context frames. At strength zero the probe is the real window, bit for bit, so its responses are the responses to reality and its AUC is 0.5 by construction; this confirms that the pipeline treats identical samples as identical, not that the test is calibrated. Two families are graded finely.

\emph{Pair swaps.} The 14 generated frames are divided into seven adjacent pairs $(1,2), (3,4), \dots, (13,14)$, and exactly $k$ of the seven pairs are swapped, $k \in \{0, 1, \dots, 7\}$. Which pairs are swapped is random per window: a fixed seed per window gives a random order of the seven pairs, and the first $k$ pairs in that order are swapped, so the swapped sets are nested in $k$ and, at each $k$, a uniformly drawn $k$-subset of the seven pairs. The set of frames is that of reality for every $k$, so the luminance control's per-component mean and standard deviation are exactly those of reality, and only its autocorrelation and mean absolute change can differ.

\emph{Slow motion by repeats.} $\mathrm{idx}(t) = 4 + \lfloor (14 - r)\, t / 14 \rfloor$, in integer arithmetic, so that exactly $r$ of the generated frames repeat their predecessor (the first of them always repeats the last context frame) and the playback speed is $v = 1 - r/14$; $r \in \{0, 1, \dots, 7, 11, 14\}$, and $r = 14$ is a freeze on the last context frame. The continuation ends at real frame $18 - r$, so the last $r$ real frames of the window do not appear: unlike the pair swaps, this family changes which frames are shown, and the luminance control's mean and standard deviation can differ from reality's.

Observers, features, classifier and folds are those of the main test (\secref{sec:measurement}): the 50 trained models, the five untrained seeds and the luminance control; the four statistics over the 29 steps of the generated frames; the logistic regression cross-validated leave-one-generator-fold-out on the same five folds. The probes involve no training and do not depend on the fold; the folds are kept so that the AUCs are comparable with \tabref{tab:main}. For each family and observer we take the curve of the AUC against strength (for the trained and the untrained observer the mean over the 50 models or the five seeds) and define the threshold at level 0.75 or 0.9 as the strength at which the curve first reaches the level, by linear interpolation between that grid point and the previous one. Its 95\% interval comes from the paired sequence-cluster bootstrap of \secref{sec:measurement}: in each of 1,000 draws the sequences are resampled, the AUC of every observer on every probe is recomputed from the stored held-out scores, the mean curves are formed again and the thresholds recomputed, with the same draws for all observers, so that differences between thresholds are paired. A draw in which a curve does not reach the level on the grid counts as lying beyond the grid, and a threshold whose interval extends beyond the grid is reported as not reached. The interval of a difference of thresholds is taken over the draws in which both curves reach the level; where one of them never does, we report the fraction of draws in which the trained curve reaches the level first.

A first, coarser run used three families on grids of four or five points: slow motion by speed, $\mathrm{idx}(t) = 4 + \lfloor v t \rfloor$ with $v \in \{0, 0.25, 0.5, 0.75, 1\}$, of which $v = 0.5$ and $v = 0$ coincide with $r = 7$ and $r = 14$; time reversal at a turning point $m$ (\emph{bounce}, $\mathrm{idx}(t) = 4 + t$ for $t \le m$ and $4 + 2m - t$ after it, $m \in \{5, 7, 10, 14\}$); and random permutation within consecutive blocks of $b$ generated frames (\emph{shuffle}, $b \in \{1, 2, 3, 5, 14\}$). Its results are in \secref{app:probes}.

\subsection{Measurement: a two-sample test on temporal response statistics}\label{sec:measurement}

The observer is simulated on every real window and on every dream. The 11 integration steps that belong to the context frames are dropped from every response, since context is identical for a dream and its real counterpart (the drop also covers the fade-in transient), leaving 29 steps per window. From the 29-step response of each of the 65 cell types we compute four statistics, the mean, the standard deviation, the lag-1 autocorrelation and the mean absolute step-to-step change, giving a 260-dimensional feature vector per window. The same four statistics computed over the 721 columns' luminance itself would ignore the brain; the \emph{direct-luminance control} instead projects each window's frames onto 65 principal components fitted on the real training frames of the fold (93--95\% of the variance), drops the context frames, and computes the same four statistics per component, so that it has the same dimension and the same temporal statistics as the brain features but no brain.

The test is a classifier two-sample test \citep{c2st}: an $\ell_2$-regularised logistic regression is trained to separate the 198 real windows from the 198 dreams of one world model at one temperature, and its held-out scores give an AUC (dream as the positive class; 0.5 is indistinguishable). The decision is the classifier's, not the observer's; the AUC measures how linearly decodable the difference between the two samples is from the observer's response statistics. Two levels of cross-validation keep sequences disjoint. The world models are trained with a five-fold cross-validation over the 35 sequences (seven held-out sequences per fold, three for early stopping, 25 for training), so every one of the 198 windows is dreamt by a generator that never saw its sequence. The classifier is then cross-validated leave-one-generator-fold-out: it is trained on the windows of four generator folds and tested on the fifth, with features standardised on the training part and the regularisation strength chosen from $\{0.01, 0.1, 1\}$ by an inner three-fold cross-validation grouped by sequence. The pooled held-out scores give the AUC; its 95\% confidence interval is a cluster bootstrap over sequences (1,000 draws), and we also report the accuracy at threshold 0.5 per fold (appendix). A one-hidden-layer MLP (64 units, early stopping) replaces the logistic regression as a robustness check. For the trained observer we report the mean, standard deviation and range of the AUC over the 50 models, for the untrained observer the mean and standard deviation over the five seeds; the confidence intervals are given for one trained model (\texttt{000}), the untrained observer with seed 0 and the luminance control. The trained-minus-untrained difference is assessed with a paired cluster bootstrap: in each of 1,000 draws the sequences are resampled with replacement, the AUC of every trained model and every untrained seed is recomputed from the stored held-out scores on the resampled windows, and the mean over trained models minus the mean over untrained seeds gives one draw of the difference.

\paragraph{Observer-seen and observer-unseen sequences.} Because the observer was trained on 17 of the sequences, we repeat the test separated by sequence. In variant (a) the classifier is trained as above on all windows and its held-out AUC is computed on the seen and on the unseen windows separately. In variant (b) the classifier is trained and tested on unseen windows only, with the same folds restricted to them. Variant (b) halves the classifier's training data, so we add its mirror image, variant (c), trained and tested on seen windows only, with a similar number of windows (96 against 102); a drop shared by (b) and (c) is an effect of training-set size, a drop in (b) alone an effect of the observer's familiarity with the sequences.

\paragraph{Density-based detectors.} For comparison with the anomaly-detection route we also fit two densities on the real training windows' features and score real and dream test windows by negative log-density (Gaussian with Ledoit--Wolf shrinkage \citep{ledoitwolf2004}) or by energy (a latent-space energy-based model, a three-layer spectral-normalised MLP of width 128 trained with short-run Langevin contrastive divergence and a replay buffer \citep{du2019implicit,lebm}, 100 epochs, no hyperparameter search); both are fitted once per fold on the fold's real train-and-validation windows and score the held-out windows, so the AUCs are directly comparable with the two-sample test.

\subsection{Detection horizon}\label{sec:horizon_setup}

To ask how quickly the observer decides, we repeat the two-sample test with features computed over only the integration steps that belong to the first $k$ generated frames, $k \in \{1, 2, 3, 4, 6, 8, 11, 14\}$, using the resampler's own frame-to-step mapping (2, 4, 6, 8, 12, 16, 23 and 29 steps). With two steps ($k=1$) the standard deviation and the autocorrelation are constants, so only the mean and the mean absolute change are used; the luminance control at $k=1$ has a single frame and uses the mean only. Everything else is as above.

\section{Results}\label{sec:results}

\begin{table}[t]
\centering
\caption{Separability of each dream from reality by the two-sample test (window AUC, dream positive; 0.5 is indistinguishable, lower is a more convincing dream), on 198 held-out windows. Trained observer: mean $\pm$ standard deviation and range over the 50 models of the ensemble; untrained observer: mean $\pm$ standard deviation over five initialisation seeds; trained $-$ untrained: difference of the two means with its 95\% paired sequence-cluster bootstrap interval; direct-luminance control: one model, with 95\% sequence-cluster bootstrap interval. The last two columns are conventional measures over the 14 generated frames: mean absolute pixel error to the real continuation (persistence error 0.108) and mean frame-to-frame change (real: 0.041).}
\label{tab:main}
\small
\resizebox{\textwidth}{!}{%
\begin{tabular}{llccccccc}
\toprule
World model & $\tau$ & Trained (50 models) & Range & Untrained (5 seeds) & Trained $-$ untrained & Luminance & $|\mathrm{real}-\mathrm{dream}|$ & Frame change \\
\midrule
Persistence walk & 0.25 & 0.96 $\pm$ 0.02 & 0.92--0.99 & 0.95 $\pm$ 0.01 & +0.01 [$-$0.01, +0.03] & 1.00 [1.00, 1.00] & 0.117 & 0.011 \\
Persistence walk & 0.5  & 0.97 $\pm$ 0.03 & 0.90--1.00 & 0.88 $\pm$ 0.02 & +0.09 [+0.04, +0.14] & 0.99 [0.97, 1.00] & 0.131 & 0.022 \\
Persistence walk & 1.0  & 0.99 $\pm$ 0.02 & 0.90--1.00 & 0.75 $\pm$ 0.00 & +0.24 [+0.18, +0.31] & 0.94 [0.90, 0.97] & 0.161 & 0.043 \\
MDN-RNN & 0.25 & 0.85 $\pm$ 0.04 & 0.75--0.96 & 0.65 $\pm$ 0.01 & +0.20 [+0.15, +0.27] & 0.67 [0.61, 0.74] & 0.144 & 0.054 \\
MDN-RNN & 0.5  & 0.89 $\pm$ 0.04 & 0.79--0.98 & 0.63 $\pm$ 0.01 & +0.26 [+0.19, +0.34] & 0.74 [0.68, 0.80] & 0.153 & 0.078 \\
MDN-RNN & 1.0  & 0.96 $\pm$ 0.03 & 0.89--1.00 & 0.58 $\pm$ 0.02 & +0.37 [+0.31, +0.44] & 0.84 [0.79, 0.89] & 0.176 & 0.115 \\
RSSM & 0.25 & 0.96 $\pm$ 0.03 & 0.79--1.00 & 0.73 $\pm$ 0.01 & +0.23 [+0.18, +0.28] & 1.00 [0.99, 1.00] & 0.128 & 0.014 \\
RSSM & 0.5  & 0.97 $\pm$ 0.03 & 0.78--0.99 & 0.69 $\pm$ 0.01 & +0.28 [+0.22, +0.33] & 1.00 [1.00, 1.00] & 0.132 & 0.019 \\
RSSM & 1.0  & 0.97 $\pm$ 0.03 & 0.81--0.99 & 0.75 $\pm$ 0.01 & +0.22 [+0.18, +0.26] & 1.00 [0.99, 1.00] & 0.151 & 0.039 \\
\bottomrule
\end{tabular}}
\end{table}

\paragraph{Every dream is linearly separable in the trained responses, and in this network training is what makes it so.} \Tabref{tab:main} gives the main result. Averaged over the 50 trained models, a logistic regression on the observer's response statistics separates real windows from every one of the nine dreams, with AUCs from 0.85 (MDN-RNN at $\tau = 0.25$) to 0.99 (persistence walk at $\tau = 1$); the smallest AUC of any single model on any dream is 0.75, and for model \texttt{000} the lowest confidence bound over the nine dreams is 0.78, so no dream is at chance in the trained response space. The MLP classifier gives the same picture (0.82--0.97 over the first five models, appendix). In the responses of the same network at its initialisation the learned world models are much less separable: 0.58--0.75 for the MDN-RNN and the RSSM (mean over five seeds; the seed-to-seed standard deviation is at most 0.02, as expected from seeds that differ only in the bias draw). The one dream that is as separable in the untrained responses as in the trained ones is the persistence walk at $\tau = 0.25$ (0.95), which is nearly a frozen frame; a static input is atypical for any dynamical system, trained or not, and the paired difference for it is the only one whose interval includes zero ($+0.01$ [$-0.01$, $+0.03$]). For the six learned-world-model dreams the paired trained-minus-untrained difference is 0.20--0.37, with every 95\% interval above 0.14, an order of magnitude larger than the spread over trained models (standard deviation 0.02--0.04).

\begin{table}[t]
\centering
\caption{Observer-seen and observer-unseen sequences (window AUC, mean $\pm$ standard deviation over the 50 trained models or the five untrained seeds; luminance control as point estimate). Seen: the 17 sequences \texttt{flyvis} was trained on (96 windows); unseen: its six validation sequences and the 12 Sintel test sequences (102 windows). (a) Classifier trained on all windows, AUC on the seen or unseen test windows. (b) Classifier trained and tested on unseen windows only. (c) Classifier trained and tested on seen windows only, the size-matched counterpart of (b). Confidence intervals are not shown; the paired trained-minus-untrained differences on unseen windows are summarised with their intervals in the text.}
\label{tab:seen}
\small
\resizebox{\textwidth}{!}{%
\begin{tabular}{llcccccccc}
\toprule
 & & \multicolumn{4}{c}{Trained (50 models)} & \multicolumn{2}{c}{Untrained (5 seeds)} & \multicolumn{2}{c}{Luminance} \\
\cmidrule(lr){3-6} \cmidrule(lr){7-8} \cmidrule(lr){9-10}
World model & $\tau$ & (a) seen & (a) unseen & (b) unseen & (c) seen & (a) unseen & (b) unseen & (a) unseen & (b) unseen \\
\midrule
Persistence walk & 0.25 & 0.96 $\pm$ 0.02 & 0.96 $\pm$ 0.02 & 0.93 $\pm$ 0.04 & 0.93 $\pm$ 0.03 & 0.95 $\pm$ 0.00 & 0.90 $\pm$ 0.01 & 1.00 & 0.98 \\
Persistence walk & 0.5  & 0.96 $\pm$ 0.03 & 0.97 $\pm$ 0.03 & 0.94 $\pm$ 0.04 & 0.92 $\pm$ 0.05 & 0.87 $\pm$ 0.02 & 0.81 $\pm$ 0.01 & 0.99 & 0.94 \\
Persistence walk & 1.0  & 0.99 $\pm$ 0.02 & 0.99 $\pm$ 0.02 & 0.94 $\pm$ 0.04 & 0.97 $\pm$ 0.04 & 0.71 $\pm$ 0.01 & 0.61 $\pm$ 0.02 & 0.95 & 0.89 \\
MDN-RNN & 0.25 & 0.84 $\pm$ 0.05 & 0.86 $\pm$ 0.05 & 0.80 $\pm$ 0.07 & 0.76 $\pm$ 0.06 & 0.68 $\pm$ 0.02 & 0.65 $\pm$ 0.02 & 0.64 & 0.59 \\
MDN-RNN & 0.5  & 0.89 $\pm$ 0.04 & 0.90 $\pm$ 0.04 & 0.83 $\pm$ 0.06 & 0.83 $\pm$ 0.05 & 0.67 $\pm$ 0.01 & 0.60 $\pm$ 0.03 & 0.71 & 0.67 \\
MDN-RNN & 1.0  & 0.95 $\pm$ 0.03 & 0.96 $\pm$ 0.03 & 0.91 $\pm$ 0.04 & 0.91 $\pm$ 0.04 & 0.59 $\pm$ 0.04 & 0.47 $\pm$ 0.01 & 0.83 & 0.82 \\
RSSM & 0.25 & 0.96 $\pm$ 0.03 & 0.97 $\pm$ 0.03 & 0.92 $\pm$ 0.05 & 0.92 $\pm$ 0.05 & 0.75 $\pm$ 0.02 & 0.72 $\pm$ 0.02 & 1.00 & 0.95 \\
RSSM & 0.5  & 0.96 $\pm$ 0.03 & 0.98 $\pm$ 0.03 & 0.92 $\pm$ 0.04 & 0.92 $\pm$ 0.04 & 0.71 $\pm$ 0.02 & 0.67 $\pm$ 0.01 & 1.00 & 0.96 \\
RSSM & 1.0  & 0.95 $\pm$ 0.03 & 0.98 $\pm$ 0.03 & 0.91 $\pm$ 0.03 & 0.92 $\pm$ 0.04 & 0.77 $\pm$ 0.01 & 0.71 $\pm$ 0.01 & 1.00 & 0.92 \\
\bottomrule
\end{tabular}}
\end{table}

\paragraph{The separation does not depend on the observer having seen the sequence.} The trained observer was fitted to 17 of our 35 sequences, so its responses could in principle separate dreams only where it recalls the statistics of its own training movies. \Tabref{tab:seen} rules this out at the resolution of our data. With the classifier trained as in the main analysis (variant a), the AUC on unseen windows (0.86--0.99) is as high as on seen windows (0.84--0.99), the difference unseen minus seen lying between $-0.003$ and $+0.029$. With the classifier trained and tested on unseen windows only (variant b), the AUC drops by 0.027--0.065 to 0.80--0.94; but the same drop occurs when the classifier is trained and tested on seen windows only (variant c, 0.76--0.97; (b) minus (c) between $-0.02$ and $+0.04$), and for the untrained observer and the luminance control as well. The drop is therefore an effect of halving the classifier's training data, not of the observer's familiarity with the sequences. On unseen windows only, the paired trained-minus-untrained difference stays at 0.15--0.44 for the learned world models, with every 95\% interval above 0.04. The observer is thus not merely recalling its training sequences' statistics.

\paragraph{The trained responses and the pixels disagree in both directions.} The direct-luminance control, the same temporal statistics on principal components of the dream's own pixels, is a different kind of judge rather than a worse one. It is near-perfect on the RSSM (AUC 0.996--0.999) and on the persistence walk (0.94--1.00): the RSSM's decoder-mean dreams are smoother than reality (spatial standard deviation 0.13--0.15 against 0.16 for real frames) and the persistence walk at low temperature is nearly static, and both are trivially separable in pixel statistics once a classifier is allowed to look for ``too smooth'' or ``too still''. On the MDN-RNN, however, the luminance control is the weakest judge (0.67--0.84), whereas the trained observer's responses reach 0.85--0.96. The MDN-RNN's defect is flicker: its frame-to-frame change (0.054--0.115) exceeds reality's (0.041) at every temperature, and it is more linearly decodable from the trained responses than from the pixel statistics.

\paragraph{The ranking is unrelated to pixel error.} We describe the rank correlations over the nine-condition grid (three world models by three temperatures), which is too small and too structured for a significance test to carry much weight. The Spearman correlation between the mean absolute pixel error and the trained observer's AUC (50-model mean) is $-0.07$: the ranking of the dreams in the trained response space has no visible relation to how close they are to the real continuation. The luminance control's ranking is anti-correlated with error ($-0.68$), and so is the untrained observer's ($-0.63$, five-seed mean): the dreams with the lowest pixel error, the persistence walk at $\tau = 0.25$ and the RSSM, are the ones their classifiers separate most easily, because low error is bought by stillness or smoothness. The dream that is hardest to separate in the trained responses is the low-temperature MDN-RNN, whose error (0.144) is mid-table; freezing and smoothing are separated regardless of their pixel error.

\subsection{Detection horizon}\label{sec:horizon}

\begin{figure}[t]
\centering
\includegraphics[width=\textwidth]{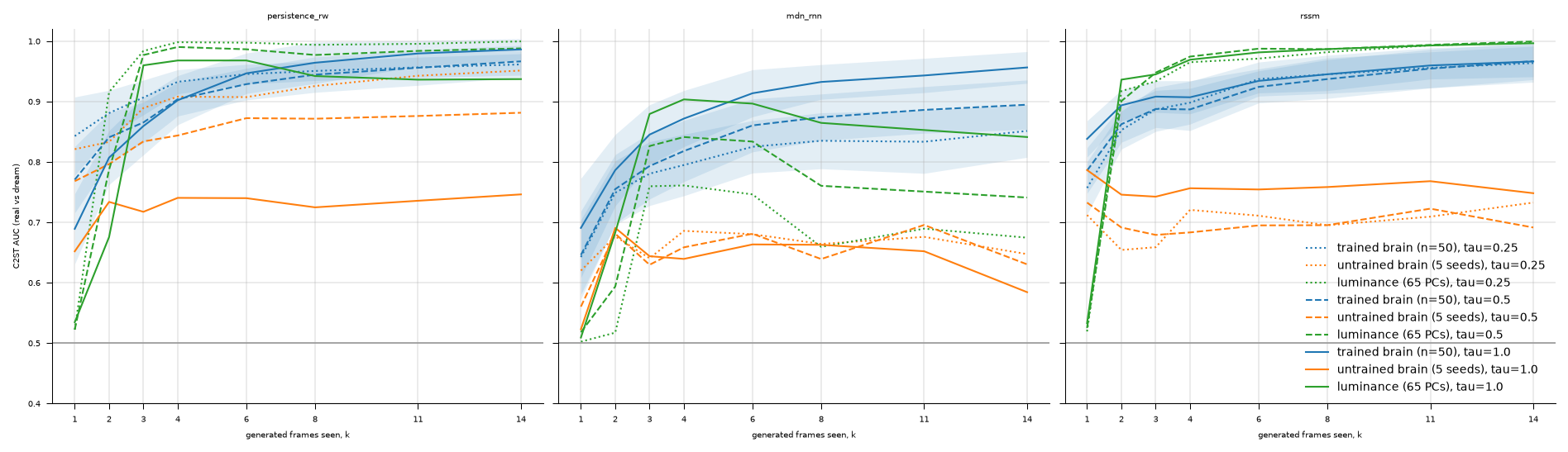}
\caption{Detection horizon: two-sample-test AUC against the number $k$ of generated frames the features are computed over, one panel per world model. Blue: trained observer, mean over the 50 models with a $\pm 1$ standard deviation band; orange: the untrained observer, mean over the five seeds; green: the direct-luminance control. Line style encodes the temperature (dotted 0.25, dashed 0.5, solid 1.0). The grey line is chance. At $k = 1$ the features use only the mean and the mean step-to-step change (two integration steps); the luminance control at $k = 1$ uses the mean of a single frame.}
\label{fig:horizon}
\end{figure}

\begin{table}[t]
\centering
\caption{Detection horizon: smallest number of generated frames $k$ at which the AUC on the trained observer's responses (mean over 50 models) exceeds 0.75 and 0.9, with the AUC at selected $k$. One generated frame is 2 integration steps (40~ms of brain time); 14 frames are 29 steps.}
\label{tab:horizon}
\small
\begin{tabular}{llcccccccc}
\toprule
World model & $\tau$ & $k > 0.75$ & $k > 0.9$ & $k=1$ & $k=2$ & $k=4$ & $k=8$ & $k=14$ \\
\midrule
Persistence walk & 0.25 & 1 & 3 & 0.84 & 0.88 & 0.93 & 0.95 & 0.96 \\
Persistence walk & 0.5  & 1 & 4 & 0.77 & 0.84 & 0.90 & 0.94 & 0.97 \\
Persistence walk & 1.0  & 2 & 4 & 0.69 & 0.81 & 0.90 & 0.96 & 0.99 \\
MDN-RNN & 0.25 & 3 & never (0.85 at 14) & 0.64 & 0.75 & 0.80 & 0.84 & 0.85 \\
MDN-RNN & 0.5  & 2 & never (0.90 at 14) & 0.65 & 0.76 & 0.82 & 0.87 & 0.90 \\
MDN-RNN & 1.0  & 2 & 6 & 0.69 & 0.79 & 0.87 & 0.93 & 0.96 \\
RSSM & 0.25 & 1 & 6 & 0.76 & 0.85 & 0.90 & 0.95 & 0.96 \\
RSSM & 0.5  & 1 & 6 & 0.79 & 0.86 & 0.89 & 0.94 & 0.97 \\
RSSM & 1.0  & 1 & 3 & 0.84 & 0.89 & 0.91 & 0.95 & 0.97 \\
\bottomrule
\end{tabular}
\end{table}

The full-length AUCs of \tabref{tab:main} saturate: the persistence walk and the RSSM sit at 0.96--0.99 and cannot be ordered. The detection horizon (\figref{fig:horizon}, \tabref{tab:horizon}) orders the world models in time instead of in level. For the persistence walk and the RSSM the AUC on the trained responses is already 0.76--0.84 after a single generated frame, two integration steps, and exceeds 0.9 by $k = 3$--6: the separation arises at the transition out of the context, where the persistence walk freezes and the RSSM's decoder mean replaces a real frame with a smoother one. The MDN-RNN is the one world model whose dreams accumulate separable evidence slowly: 0.64--0.69 at $k = 1$, 0.75--0.79 at $k = 2$, and at $\tau = 0.25$ and $\tau = 0.5$ the AUC never reaches 0.9 within the 14 generated frames (0.85 and 0.90 at $k = 14$); only at $\tau = 1$ does it cross 0.9, at $k = 6$. On the observer's clock, the low-temperature MDN-RNN is the most convincing of the nine dreams, and by a wide margin at short horizons (0.75 against 0.85--0.88 at $k = 2$, 0.80 against 0.90 at $k = 4$), a difference the full-length numbers hid. The luminance control reads the persistence walk and the RSSM within two or three frames (0.90--0.98), but on the MDN-RNN its AUC peaks at $k = 4$--6 (0.76--0.90) and then falls (0.67--0.84 at $k = 14$): the MDN-RNN's pixel statistics regress towards typical values over time while the trained responses keep accumulating separable evidence. The untrained observer (mean over the five seeds, whose standard deviation is at most 0.02) does not build up separation in the same way: on the learned world models it stays between 0.58 and 0.77 from $k = 2$ to $k = 14$ (0.52--0.79 at $k = 1$), whereas the trained observer rises from 0.75--0.89 at $k = 2$ to 0.85--0.97 at $k = 14$, so the shape of the horizon curve is a property of the trained dynamics.

\subsection{Graded temporal probes}\label{sec:probes}

\begin{figure}[t]
\centering
\includegraphics[width=\textwidth]{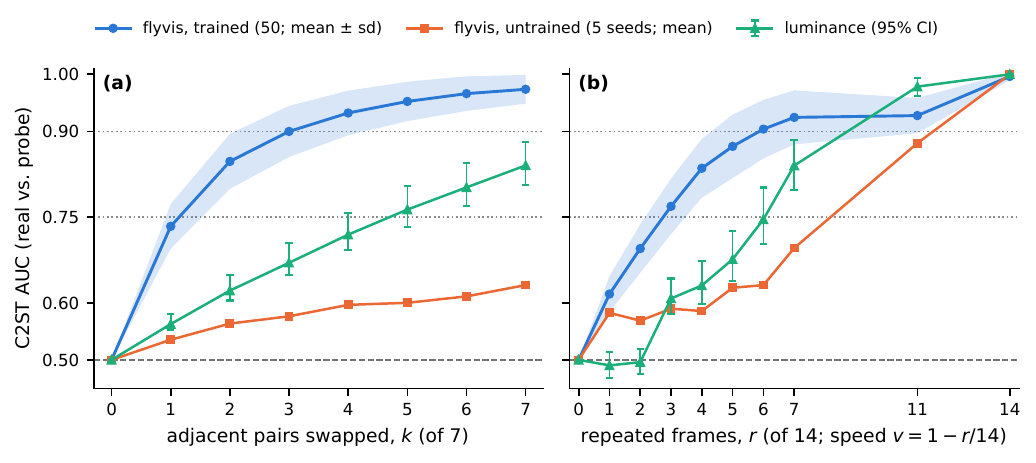}
\caption{Graded temporal probes: two-sample-test AUC against the strength of the probe. (a) Number $k$ of the seven adjacent pairs of generated frames that are swapped; (b) number $r$ of the 14 generated frames that repeat their predecessor in slow motion (speed $v = 1 - r/14$; $r = 14$ is a freeze on the last context frame). Blue: trained observer, mean over the 50 models with a $\pm 1$ standard deviation band; orange: untrained observer, mean over the five seeds; green: direct-luminance control with its 95\% sequence-cluster bootstrap interval. Dotted lines mark the threshold levels 0.75 and 0.9, the dashed line chance. Strength zero is reality itself, where the AUC is 0.5 by construction.}
\label{fig:probes}
\end{figure}

\begin{table}[t]
\centering
\caption{Thresholds of the graded probes: the strength ($k$ swapped pairs of seven, or $r$ repeated frames of 14) at which the AUC curve first reaches 0.75 or 0.9, by linear interpolation on the grid, with 95\% paired sequence-cluster bootstrap intervals (1,000 draws). Trained: mean curve over the 50 models; untrained: mean curve over the five seeds. The last two columns are paired differences of thresholds; a positive value means that the trained curve reaches the level at a weaker probe. Not reached: the curve stays below the level on the whole grid; for a difference involving such a curve we give the fraction of draws in which the trained curve reaches the level first.}
\label{tab:thresholds}
\small
\resizebox{\textwidth}{!}{%
\begin{tabular}{llccccc}
\toprule
Probe & Level & Trained (50 models) & Untrained (5 seeds) & Luminance & Untrained $-$ trained & Luminance $-$ trained \\
\midrule
Pair swaps, $k$ & 0.75 & 1.1 [0.9, 1.4] & not reached (0.63 at $k = 7$) & 4.7 [3.9, 5.4] & trained first in 100\% & +3.6 [+2.8, +4.3] \\
Pair swaps, $k$ & 0.9  & 3.0 [2.4, 3.8] & not reached & not reached (0.84 at $k = 7$) & trained first in 100\% & trained first in 100\% \\
Slow motion, $r$ & 0.75 & 2.7 [2.3, 3.2] & 8.2 [7.2, 8.9] & 6.0 [5.3, 6.5] & +5.4 [+4.6, +6.1] & +3.3 [+2.6, +3.8] \\
Slow motion, $r$ & 0.9  & 5.9 [4.8, 8.3] & 11.5 [10.8, 12.0] & 8.7 [7.6, 9.4] & +5.6 [+3.7, +6.6] & +2.9 [+0.9, +3.8] \\
\bottomrule
\end{tabular}}
\end{table}

\paragraph{One swapped pair is linearly decodable from the trained responses.} Swapping a single adjacent pair of the 14 generated frames, an exchange of two real frames that leaves the set of frames and every per-frame statistic as they are in reality, gives an AUC of 0.73 $\pm$ 0.04 on the trained responses (mean over the 50 models; \figref{fig:probes}a), against 0.54 for the untrained network and 0.56 [0.55, 0.58] for the luminance control; the paired differences are $+0.20$ [$+0.16$, $+0.24$] and $+0.17$ [$+0.14$, $+0.20$]. The trained curve rises steeply, to 0.85 at $k = 2$, 0.90 at $k = 3$ and 0.97 when all seven pairs are swapped. The luminance control, which here can only use the autocorrelation and the mean absolute change of its components, rises roughly linearly to 0.84 at $k = 7$; the untrained network stays between 0.54 and 0.63 over the whole grid. From $k = 2$ on, the paired difference is at least $+0.28$ to the untrained network and at least $+0.13$ to the luminance control, with every interval above $+0.09$ (\tabref{tab:probes_fine} in the appendix).

\paragraph{Slow motion is separable from a single repeated frame.} With one repeated frame of 14 ($r = 1$, speed 13/14) the trained observer gives 0.62 $\pm$ 0.03, and every one of the 50 models at least 0.57 (\figref{fig:probes}b). The luminance control is at chance at $r = 1$ and $r = 2$ (0.49 [0.47, 0.51] and 0.50 [0.48, 0.52]); the untrained network is above chance at $r = 1$ (0.58), so that the paired difference there is small ($+0.03$ [$+0.01$, $+0.05$]), but it stays nearly flat over $r = 1$--4 (0.57--0.59) while the trained curve rises to 0.77 at $r = 3$, 0.84 at $r = 4$ and 0.90 at $r = 6$. Beyond that the trained curve nearly stops, 0.92 at $r = 7$ and 0.93 at $r = 11$, before the freeze ($r = 14$, 1.00). Over $r = 2$--6 the trained observer exceeds the luminance control by $+0.16$ to $+0.21$; at $r = 11$, where the continuation omits the last 11 real frames and the pixel statistics depart most from reality's, the luminance control is the better judge (0.98 against 0.93, difference $-0.05$ [$-0.08$, $-0.03$]).

\paragraph{The trained responses reach each level at a weaker probe.} \Tabref{tab:thresholds} gives the thresholds. For pair swaps the trained curve reaches 0.75 at $k = 1.1$ [0.9, 1.4] and the luminance control at 4.7 [3.9, 5.4], a paired difference of $+3.6$ [$+2.8$, $+4.3$] pairs; the untrained curve does not reach 0.75 on the grid, nor in any bootstrap draw. The trained curve reaches 0.9 at $k = 3.0$ [2.4, 3.8], and neither control reaches it (the luminance control in 0.5\% of the draws). For slow motion the thresholds at 0.75 are $r = 2.7$ [2.3, 3.2] for the trained observer, 6.0 [5.3, 6.5] for the luminance control and 8.2 [7.2, 8.9] for the untrained network, and at 0.9 they are 5.9 [4.8, 8.3], 8.7 [7.6, 9.4] and 11.5 [10.8, 12.0]. The trained threshold is the smaller in every bootstrap draw for each comparison except slow motion at 0.9 against the luminance control, where it is the smaller in 97.6\% of the draws ($+2.9$ [$+0.9$, $+3.8$]). The slow-motion thresholds at 0.9 need care. The trained interval is long on its upper side because the trained curve is nearly flat between $r = 6$ (0.90) and $r = 11$ (0.93): in a draw in which it stays just below 0.9 at $r = 6$ and 7, the crossing moves into the interval between $r = 7$ and 11. The untrained threshold at 0.75 and the luminance threshold at 0.9 lie in that same gap of four repeats, and the untrained threshold at 0.9 between $r = 11$ and 14, so they are interpolations over long steps of the grid. The curves need not be monotone (the untrained slow-motion curve is 0.58 at $r = 1$ and 0.57 at $r = 2$), and the thresholds are first crossings.

\paragraph{The spread over trained models.} Computed for each trained model separately, the threshold at 0.75 has a median of $k = 1.1$ (interquartile range 1.0--1.3) for pair swaps; 49 of the 50 models reach 0.75 by $k = 2$, and the slowest at $k = 4.9$. At 0.9 the median is $k = 2.8$ (2.4--3.3), and one model does not reach 0.9 on the grid. For slow motion the median threshold is $r = 2.8$ (2.5--3.0) at 0.75, with 49 of the 50 models at $r \le 4$, and $r = 5.7$ (4.7--7.0) at 0.9, with a range of 3.8--12.3 that reflects the flat stretch of the curve. The spread over models is wide, but at 0.75 the per-model thresholds lie, with one exception each, below the thresholds of both controls.

\paragraph{No advantage on the observer-seen sequences.} Repeating the seen/unseen analysis of \secref{sec:results} for pair swaps, the trained observer's AUC on unseen windows (variant a) is higher than on seen windows by 0.004--0.03; the luminance control shows the same direction by more (0.02--0.07), and the untrained network differs by less than 0.01. Trained and tested on unseen windows only (variant b), the trained observer is 0.01--0.05 higher than when trained and tested on seen windows only (variant c). For slow motion the unseen-minus-seen difference of the trained observer lies between $-0.03$ and $+0.02$. The separation of the probes is thus not larger on the sequences the observer was trained on; the small differences that exist go the other way and are shared with the pixel statistics, so they reflect the content of the sequences rather than the observer's familiarity with them.

\subsection{What the classifier uses}\label{sec:attribution}

As a descriptive account of where the separation lies, we fitted the logistic regression on all 396 windows of model \texttt{000} at $\tau = 0.5$ and summed the absolute standardised coefficients per statistic and per cell type. The lag-1 autocorrelation carries the largest share of the weight for every world model (0.41 for the persistence walk, 0.36 for the MDN-RNN, 0.29 for the RSSM, against 0.14--0.28 for the mean, 0.18--0.21 for the standard deviation and 0.21--0.28 for the mean absolute change). The ten cell types with the largest weight are, for the persistence walk, Tm4, TmY15, TmY14, T5b, TmY5a, T4b, T5d, Mi2, T4c and Tm30; for the MDN-RNN, Tm5a, Tm4, TmY13, TmY4, TmY5a, Mi3, TmY14, T5c, Mi12 and Mi15; for the RSSM, Tm4, TmY4, TmY15, Mi12, T5b, TmY14, Mi3, T5d, Tm1 and T5c; each of these carries between 2\% and 7\% of the total weight. The diagnostic quantity is thus the temporal correlation of responses in the medulla-to-lobula pathway, with the same types recurring across world models; we report this as a description of the classifier, not as a mechanistic claim about the fly.

\subsection{One-class density scores}\label{sec:density}

We also tried the anomaly-detection route: fit a density on the observer's responses to real training windows and score a dream by its negative log-density (shrinkage Gaussian) or energy (LEBM). On the same features these one-class scores reach only 0.56--0.69 (models \texttt{000}--\texttt{004}; \tabref{tab:density} in the appendix), and in several cells they fall below 0.5, that is, the dreams are more typical than reality under the fitted density: the luminance LEBM on the RSSM at $\tau = 0.25$ (0.40, interval [0.28, 0.52]), a per-step Gaussian on the RSSM (0.26--0.39) and the untrained observer's Gaussian on the learned world models (0.40--0.47). One-class typicality scores need not align with two-sample distinguishability, especially for temporally simplified continuations: a smooth dream can sit in a high-density region of a density fitted on real responses, much as simple images receive high likelihood from generative models trained on complex ones \citep{nalisnick2019}, whereas the two-sample test sees both samples and is symmetric in them. The two methods also serve different purposes: the two-sample test is fitted per dream source with real and dream labels, while a one-class detector has to flag deviations it has never seen. The LEBM's weakness may in addition reflect the data size, fewer than 200 real windows per fold in 260 dimensions, and hyperparameters that we did not tune. We therefore report the density scores as a secondary observation about this setting, not as evidence against energy-based detectors in general.

\subsection{The ensemble is fifty draws of one judge}\label{sec:ensemble}

Over the 50 trained models the standard deviation of the AUC is 0.02--0.04 per dream at full length (up to 0.08 at $k = 1$), small against the differences between dreams and between observers. The per-model AUC is only weakly related to the model's own quality on its training task: the Spearman correlation between a model's mean AUC over the nine dreams and its \texttt{flyvis} validation loss (the optic-flow loss of the stored checkpoint, 5.14--5.68 across the ensemble) is $-0.17$, and per dream it ranges from $-0.32$ to $-0.04$, with the same sign throughout (models with a lower flow loss give slightly higher AUCs). We therefore treat the ensemble as 50 draws of the same observer rather than as 50 observers of different quality, and report its mean and spread. The 50 models share the connectome, the training data and the training procedure, so they are not 50 independent replications of the measurement; the uncertainty that matters for the comparison with the untrained observer is the sequence-level one, which the paired cluster bootstrap captures.

\section{Discussion}\label{sec:discussion}

\paragraph{What the trained responses tolerate and what they expose.} Read through the trained observer's responses, a dream is hard to separate from reality to the extent that its temporal statistics, above all the autocorrelation of cell-type responses, match those of reality. The low-temperature MDN-RNN, a dream with mediocre pixel error but with approximately the right amount of change from frame to frame, is separated only gradually; a frozen frame and a smoothed continuation are separated at once, however small their pixel error. This is the opposite of what a pixel-error ranking says, and also of what pixel statistics say, which reward stillness and smoothness. The observer's response space thus provides an evaluation axis that is not the pixel axis, and the detection horizon turns it into a graded score where the full-length test saturates. The graded probes show that this axis is fine-grained. A continuation in which every frame is real and only two neighbouring frames have changed places, or one frame is shown twice, is linearly decodable from the trained responses at strengths at which the pixel statistics are at or near chance and the untrained network barely departs from it; the trained responses thus expose local violations of temporal order, not only the gross defects of small world models. They do not expose every temporal defect better: a reversal of time at a turning point is as separable in the untrained responses as in the trained ones, or more (\secref{app:probes}), and in strong slow motion, where many real frames are missing, the pixel statistics are the better judge.

\paragraph{What ``can tell'' means here.} The answer to whether the brain can tell is given in a precise and limited form. The observer does not decide anything; a logistic regression trained on its response statistics does, and our finding is that the trained observer's response space makes the temporal defects of these rollouts linearly decodable, more so than the same network's response space before training and, for flicker-type defects, more so than the pixel statistics. The probes extend this to local violations of temporal order: a swapped pair or a repeated frame is linearly decodable from the trained responses at strengths where neither control separates it as well. The thresholds of \tabref{tab:thresholds} are the strengths at which a linear classifier on response statistics reaches a given AUC; the resemblance of the curves to psychometric functions is one of form, and they are not perceptual thresholds of the fly. Whether the fly would behave differently in the dream is not addressed: the loop is open, and no readout of the fly's own behaviour is involved.

\paragraph{Limitations.} The study is confined to one visual domain, Sintel luminance on a 721-column hexagonal lattice, which is the domain the observer was trained on; the separation is the same on the sequences the observer never saw (\tabref{tab:seen}), but the visual statistics of those sequences are still Sintel's. The dreams are low-dimensional and short (14 generated frames, 0.58~s), so long-horizon drift is not probed. The generators are small and trained on about 3,400 frames per fold, and the RSSM emits its decoder mean, so its dreams carry no observation noise and are smoother than a sampled decoder would make them; a stronger generator would make the dreams harder, and the horizon curve rather than the full-length AUC is where its progress would show. We did not test large pretrained video generators: they continue the context in pixel space through their own autoencoders, so a fair test needs, besides the dream, a null in which the real frames are only passed through the same autoencoder, and that comparison is left to future work. The question is what the responses of a model of the fly's visual system make available, not which representation detects dreams best, so we do not report comparisons with generic learned representations such as motion energy, optical-flow statistics or a pretrained video encoder; such representations may well separate the dreams as well or better, and our results do not imply that the fly's visual system is a better detector than they are. Our controls are the untrained connectome and the pixel statistics, so the specific contribution of the connectome and of \texttt{flyvis} within the model is not isolated. The five untrained seeds differ only in the draw of the cell-type biases, so they span a narrow part of the space of untrained networks; that the untrained network separates the learned world models much less well than the trained one is a property of this initialisation, and a randomly initialised network of another architecture need not share it. The observer does not act, so the question of whether acting in the dream reveals it faster (the observer's outputs steering the generator) is left open as well. The graded probes are built from real frames and are not world-model dreams: they measure how much temporal disorder the response space needs before a classifier can use it, not how a generator fails. Each of the four statistics is unchanged when a response series is reversed in time, so the probes do not measure sensitivity to the direction of time, and a reversal is visible to the features mainly at its turning point and through the frames it repeats and omits. The thresholds rest on a discrete grid with linear interpolation, which is coarse between $r = 7$ and $r = 14$; the swapped pairs are drawn once per window, so the variation over which pairs are swapped is not part of the intervals. The attribution to cell types and the probe thresholds are descriptions of a linear classifier, not an account of the fly's visual system.

\paragraph{Outlook.} Three directions follow. Other brains: models of the fly visual system are an active line of work, with complete connectomes of the optic lobe \citep{nern2025} and of the male central nervous system \citep{malecns} now available and the released \texttt{flyvis} ensemble itself under scrutiny \citep{flyhack,zhou2026geometry}; as connectome-constrained models built on these data are released and validated, the same protocol can ask whether a more complete brain carries more of the dreams' deviations, and which parts of it carry them. The closed loop: a fixed, textbook readout of the observer's motion-detector responses (for instance the left--right difference of the T4/T5 populations) can drive the viewpoint of the generator, and the detection horizon can then be measured as a function of interaction time, which is the quantitative form of ``the longer one acts in the dream, the clearer it is a dream''. The benchmark: the protocol scores any world model that can continue a 721-column luminance sequence, with an observer that is public, fixed and biologically constrained; at full length the scale is saturated by current small generators, but the horizon curve has range, and we propose the smallest $k$ at which the AUC on a fixed observer's responses exceeds a given value as an observer-referenced measure of a world model's temporal fidelity; it measures fidelity to reality, not usefulness for control, which, as the temperature of \citet{ha2018} shows, can call for a less faithful dream.

\section{Reproducibility and data statement}\label{sec:repro}

Code and result files are not released with this preprint. All tables and figures are generated from stored result files, in which the random seeds, fold assignments and model configurations are recorded together with the held-out classifier scores of every window, observer and dream, so that the bootstrap intervals and the probe thresholds are recomputed from these scores without refitting; the probes' per-window seeds and frame index maps are recorded with them. The observer is \texttt{flyvis} 1.2.0 \citep{flyvis}, MIT licence, with its released pretrained ensemble; we did not retrain or modify it. MPI-Sintel \citep{sintel} is used for academic purposes only, as its distribution requests, and no frames are reproduced here. All computation ran on the CPU of a single Mac Studio (Apple M2 Ultra): simulating the observer on the 198 real windows takes about 80~s per model and on the dreams of one world model about 55~s per model and fold; training a world model takes 121--143~s per fold; the two-sample tests over all observers, world models and temperatures take about two minutes, the horizon sweep over 50 models about 15 minutes; extending the observer from five to all 50 models took 1~h~42~min of wall time with up to 16 parallel simulations, the four additional untrained seeds 9~min with 16 threads, and the seen/unseen analysis with the paired bootstrap under one minute. The temporal probes need no training; simulating all 55 observers on the coarse probes took 69~min of wall time, and the fine run, simulation and scoring together, 90~min, each with 16 processes of two threads on the Mac Studio's CPU.

\bibliography{../flydream}
\bibliographystyle{tmlr}

\appendix

\section{Additional results}\label{app:results}

\subsection{Central-cell features}\label{app:central}

\Tabref{tab:central} repeats the two-sample test with the response of the cell in the central column of each type instead of the type-mean. The picture is the same, with the untrained observer (seed 0) somewhat stronger on this variant.

\begin{table}[h]
\centering
\caption{Two-sample test on central-cell features (260 dimensions). Brain \texttt{000} with 95\% sequence-cluster bootstrap intervals for the logistic regression and the MLP; mean $\pm$ standard deviation over models \texttt{000}--\texttt{004}; untrained observer (seed 0).}
\label{tab:central}
\small
\begin{tabular}{llcccc}
\toprule
World model & $\tau$ & Brain 000, LR & Brain 000, MLP & LR, 5 brains & Untrained, LR \\
\midrule
Persistence walk & 0.25 & 0.96 [0.94, 0.98] & 0.94 [0.90, 0.97] & 0.92 $\pm$ 0.03 & 0.91 [0.87, 0.94] \\
Persistence walk & 0.5  & 0.96 [0.93, 0.97] & 0.93 [0.90, 0.96] & 0.91 $\pm$ 0.04 & 0.87 [0.82, 0.91] \\
Persistence walk & 1.0  & 0.97 [0.95, 0.98] & 0.97 [0.95, 0.99] & 0.94 $\pm$ 0.02 & 0.88 [0.84, 0.91] \\
MDN-RNN & 0.25 & 0.90 [0.87, 0.93] & 0.87 [0.83, 0.92] & 0.89 $\pm$ 0.02 & 0.71 [0.65, 0.76] \\
MDN-RNN & 0.5  & 0.93 [0.90, 0.96] & 0.88 [0.82, 0.93] & 0.92 $\pm$ 0.02 & 0.78 [0.74, 0.82] \\
MDN-RNN & 1.0  & 0.96 [0.94, 0.98] & 0.95 [0.92, 0.97] & 0.95 $\pm$ 0.02 & 0.86 [0.82, 0.89] \\
RSSM & 0.25 & 0.97 [0.95, 0.98] & 0.95 [0.91, 0.98] & 0.96 $\pm$ 0.01 & 0.85 [0.79, 0.90] \\
RSSM & 0.5  & 0.97 [0.95, 0.98] & 0.97 [0.95, 0.98] & 0.96 $\pm$ 0.01 & 0.80 [0.73, 0.87] \\
RSSM & 1.0  & 0.96 [0.93, 0.98] & 0.95 [0.92, 0.98] & 0.94 $\pm$ 0.01 & 0.71 [0.66, 0.76] \\
\bottomrule
\end{tabular}
\end{table}

\subsection{Per-fold accuracies}\label{app:folds}

\Tabref{tab:folds} gives the two-sample-test accuracy at threshold 0.5 for each classifier test fold (one generator fold held out at a time), for model \texttt{000}, the untrained observer (seed 0) and the luminance control. The MLP robustness check gives, over models \texttt{000}--\texttt{004}, 0.94 $\pm$ 0.01, 0.93 $\pm$ 0.04 and 0.97 $\pm$ 0.03 for the persistence walk, 0.82 $\pm$ 0.03, 0.88 $\pm$ 0.04 and 0.94 $\pm$ 0.03 for the MDN-RNN, and 0.94 $\pm$ 0.02, 0.95 $\pm$ 0.02 and 0.94 $\pm$ 0.01 for the RSSM, at $\tau = 0.25$, 0.5 and 1.0 respectively.

\begin{table}[h]
\centering
\caption{Accuracy at threshold 0.5 per classifier test fold (mean; folds 0--4), logistic regression on type-mean features (brains) and principal-component features (luminance).}
\label{tab:folds}
\small
\resizebox{\textwidth}{!}{%
\begin{tabular}{llccc}
\toprule
World model & $\tau$ & Brain 000 & Untrained & Luminance \\
\midrule
Persistence walk & 0.25 & 0.92; 0.96, 0.88, 0.88, 0.95, 0.90 & 0.90; 1.00, 0.88, 0.87, 0.90, 0.83 & 0.97; 1.00, 0.92, 0.96, 1.00, 0.96 \\
Persistence walk & 0.5  & 0.88; 0.90, 0.86, 0.86, 0.90, 0.88 & 0.79; 0.97, 0.71, 0.76, 0.79, 0.72 & 0.93; 1.00, 0.85, 0.90, 1.00, 0.92 \\
Persistence walk & 1.0  & 0.94; 0.90, 1.00, 0.94, 0.93, 0.94 & 0.69; 0.78, 0.55, 0.67, 0.71, 0.75 & 0.86; 0.94, 0.82, 0.82, 0.87, 0.83 \\
MDN-RNN & 0.25 & 0.76; 0.93, 0.64, 0.74, 0.80, 0.69 & 0.60; 0.78, 0.40, 0.68, 0.58, 0.54 & 0.57; 0.53, 0.48, 0.60, 0.63, 0.64 \\
MDN-RNN & 0.5  & 0.82; 0.94, 0.65, 0.76, 0.90, 0.83 & 0.62; 0.75, 0.52, 0.68, 0.60, 0.57 & 0.66; 0.64, 0.60, 0.68, 0.70, 0.67 \\
MDN-RNN & 1.0  & 0.93; 0.93, 0.90, 0.95, 0.98, 0.89 & 0.56; 0.62, 0.39, 0.61, 0.58, 0.60 & 0.77; 0.85, 0.82, 0.67, 0.75, 0.76 \\
RSSM & 0.25 & 0.92; 0.92, 0.93, 0.92, 0.94, 0.90 & 0.67; 0.65, 0.71, 0.64, 0.68, 0.64 & 0.96; 0.99, 0.96, 0.94, 0.99, 0.93 \\
RSSM & 0.5  & 0.92; 0.92, 0.96, 0.93, 0.92, 0.88 & 0.65; 0.60, 0.74, 0.60, 0.68, 0.62 & 0.98; 0.99, 0.99, 0.98, 1.00, 0.96 \\
RSSM & 1.0  & 0.93; 0.94, 0.93, 0.92, 0.96, 0.88 & 0.67; 0.65, 0.74, 0.64, 0.71, 0.61 & 0.99; 0.96, 1.00, 0.99, 1.00, 0.99 \\
\bottomrule
\end{tabular}}
\end{table}

\subsection{Density-based detectors}\label{app:density}

\begin{table}[h]
\centering
\caption{Two-sample test against density-based detectors on the same 260-dimensional type-mean features (models \texttt{000}--\texttt{004}, mean $\pm$ standard deviation; held-out AUC as in \tabref{tab:main}). Density detectors are fitted on the real training windows of each fold; the latent-space EBM (LEBM) scores by energy, the Gaussian by negative log-density. Values below 0.5 mean that dreams receive higher density than real windows. The last column is the luminance control's LEBM, with its 95\% interval.}
\label{tab:density}
\small
\begin{tabular}{llcccc}
\toprule
World model & $\tau$ & Two-sample test & LEBM & Gaussian & LEBM on luminance \\
\midrule
Persistence walk & 0.25 & 0.97 $\pm$ 0.02 & 0.61 $\pm$ 0.05 & 0.69 $\pm$ 0.12 & 0.56 [0.46, 0.66] \\
Persistence walk & 0.5  & 0.97 $\pm$ 0.03 & 0.62 $\pm$ 0.05 & 0.67 $\pm$ 0.06 & 0.63 [0.54, 0.72] \\
Persistence walk & 1.0  & 0.99 $\pm$ 0.01 & 0.68 $\pm$ 0.02 & 0.69 $\pm$ 0.08 & 0.77 [0.70, 0.83] \\
MDN-RNN & 0.25 & 0.86 $\pm$ 0.08 & 0.64 $\pm$ 0.06 & 0.56 $\pm$ 0.04 & 0.77 [0.70, 0.83] \\
MDN-RNN & 0.5  & 0.92 $\pm$ 0.05 & 0.67 $\pm$ 0.07 & 0.60 $\pm$ 0.03 & 0.79 [0.72, 0.84] \\
MDN-RNN & 1.0  & 0.98 $\pm$ 0.02 & 0.65 $\pm$ 0.09 & 0.69 $\pm$ 0.06 & 0.72 [0.63, 0.79] \\
RSSM & 0.25 & 0.97 $\pm$ 0.02 & 0.63 $\pm$ 0.10 & 0.58 $\pm$ 0.06 & 0.40 [0.28, 0.52] \\
RSSM & 0.5  & 0.97 $\pm$ 0.02 & 0.63 $\pm$ 0.09 & 0.56 $\pm$ 0.06 & 0.54 [0.43, 0.65] \\
RSSM & 1.0  & 0.97 $\pm$ 0.01 & 0.58 $\pm$ 0.06 & 0.57 $\pm$ 0.06 & 0.70 [0.63, 0.77] \\
\bottomrule
\end{tabular}
\end{table}

\Tabref{tab:density_full} gives the density-based AUCs for model \texttt{000} with intervals, for the untrained observer (seed 0) and for the luminance control, together with the per-step Gaussian (a Gaussian on the 65-dimensional per-step response, the window scored by its mean negative log-density), which is the detector that falls furthest below 0.5 on the RSSM.

\begin{table}[h]
\centering
\caption{Density-based detectors (held-out AUC). LEBM and Gaussian on type-mean window features for model \texttt{000} (with 95\% intervals for the LEBM) and the untrained observer (seed 0); per-step Gaussian for brain \texttt{000}; LEBM and Gaussian on principal-component features for the luminance control.}
\label{tab:density_full}
\small
\resizebox{\textwidth}{!}{%
\begin{tabular}{llccccccc}
\toprule
World model & $\tau$ & LEBM 000 & Gauss. 000 & Step Gauss. 000 & LEBM untrained & Gauss. untrained & LEBM lum. & Gauss. lum. \\
\midrule
Persistence walk & 0.25 & 0.62 [0.55, 0.70] & 0.84 & 0.37 & 0.70 [0.65, 0.76] & 0.86 & 0.56 [0.46, 0.66] & 0.60 \\
Persistence walk & 0.5  & 0.63 [0.55, 0.71] & 0.76 & 0.49 & 0.59 [0.53, 0.65] & 0.72 & 0.63 [0.54, 0.72] & 0.60 \\
Persistence walk & 1.0  & 0.66 [0.59, 0.73] & 0.70 & 0.72 & 0.58 [0.51, 0.65] & 0.49 & 0.77 [0.70, 0.83] & 0.61 \\
MDN-RNN & 0.25 & 0.68 [0.60, 0.75] & 0.53 & 0.49 & 0.58 [0.51, 0.64] & 0.47 & 0.77 [0.70, 0.83] & 0.56 \\
MDN-RNN & 0.5  & 0.75 [0.68, 0.81] & 0.58 & 0.60 & 0.57 [0.51, 0.64] & 0.44 & 0.79 [0.72, 0.84] & 0.57 \\
MDN-RNN & 1.0  & 0.77 [0.71, 0.83] & 0.69 & 0.76 & 0.59 [0.53, 0.66] & 0.42 & 0.72 [0.63, 0.79] & 0.60 \\
RSSM & 0.25 & 0.73 [0.67, 0.79] & 0.59 & 0.26 & 0.58 [0.52, 0.64] & 0.44 & 0.40 [0.28, 0.52] & 0.29 \\
RSSM & 0.5  & 0.74 [0.68, 0.80] & 0.54 & 0.29 & 0.56 [0.51, 0.62] & 0.40 & 0.54 [0.43, 0.65] & 0.39 \\
RSSM & 1.0  & 0.66 [0.60, 0.74] & 0.56 & 0.39 & 0.53 [0.48, 0.59] & 0.40 & 0.70 [0.63, 0.77] & 0.54 \\
\bottomrule
\end{tabular}}
\end{table}

\subsection{Temporal probes: the full grid and the coarse run}\label{app:probes}

\Tabref{tab:probes_fine} gives the AUCs behind \figref{fig:probes}. \Tabref{tab:probes_coarse} reports every probe of the first, coarser run (\secref{sec:probes_setup}), with the same observers, features and test.

\begin{table}[h]
\centering
\caption{Graded temporal probes (window AUC, probe positive, 198 held-out windows). Pair swaps: $k$ of seven adjacent pairs of generated frames swapped; slow motion: $r$ of 14 generated frames repeated (speed $v = 1 - r/14$). Columns as in \tabref{tab:main}: trained observer mean $\pm$ standard deviation and range over the 50 models; untrained observer mean $\pm$ standard deviation over five seeds; luminance control with its 95\% sequence-cluster bootstrap interval; differences with 95\% paired sequence-cluster bootstrap intervals. Strength zero is reality itself (AUC 0.5 for every observer by construction) and is omitted.}
\label{tab:probes_fine}
\small
\resizebox{\textwidth}{!}{%
\begin{tabular}{llcccccc}
\toprule
Probe & Strength & Trained (50 models) & Range & Untrained (5 seeds) & Trained $-$ untrained & Luminance & Trained $-$ luminance \\
\midrule
Pair swaps & $k = 1$ & 0.73 $\pm$ 0.04 & 0.56--0.79 & 0.54 $\pm$ 0.01 & +0.20 [+0.16, +0.24] & 0.56 [0.55, 0.58] & +0.17 [+0.14, +0.20] \\
Pair swaps & $k = 2$ & 0.85 $\pm$ 0.05 & 0.61--0.90 & 0.56 $\pm$ 0.01 & +0.28 [+0.25, +0.32] & 0.62 [0.60, 0.65] & +0.23 [+0.19, +0.26] \\
Pair swaps & $k = 3$ & 0.90 $\pm$ 0.04 & 0.66--0.95 & 0.58 $\pm$ 0.01 & +0.32 [+0.28, +0.36] & 0.67 [0.65, 0.70] & +0.23 [+0.19, +0.26] \\
Pair swaps & $k = 4$ & 0.93 $\pm$ 0.04 & 0.71--0.98 & 0.60 $\pm$ 0.00 & +0.34 [+0.30, +0.37] & 0.72 [0.69, 0.76] & +0.21 [+0.17, +0.24] \\
Pair swaps & $k = 5$ & 0.95 $\pm$ 0.03 & 0.76--0.99 & 0.60 $\pm$ 0.01 & +0.35 [+0.32, +0.38] & 0.76 [0.73, 0.80] & +0.19 [+0.15, +0.22] \\
Pair swaps & $k = 6$ & 0.97 $\pm$ 0.03 & 0.79--0.99 & 0.61 $\pm$ 0.01 & +0.35 [+0.32, +0.38] & 0.80 [0.77, 0.84] & +0.16 [+0.12, +0.20] \\
Pair swaps & $k = 7$ & 0.97 $\pm$ 0.03 & 0.83--0.99 & 0.63 $\pm$ 0.01 & +0.34 [+0.31, +0.38] & 0.84 [0.81, 0.88] & +0.13 [+0.09, +0.17] \\
\midrule
Slow motion & $r = 1$ & 0.62 $\pm$ 0.03 & 0.57--0.71 & 0.58 $\pm$ 0.01 & +0.03 [+0.01, +0.05] & 0.49 [0.47, 0.51] & +0.12 [+0.10, +0.15] \\
Slow motion & $r = 2$ & 0.69 $\pm$ 0.04 & 0.58--0.79 & 0.57 $\pm$ 0.01 & +0.13 [+0.09, +0.16] & 0.50 [0.48, 0.52] & +0.20 [+0.17, +0.23] \\
Slow motion & $r = 3$ & 0.77 $\pm$ 0.05 & 0.57--0.87 & 0.59 $\pm$ 0.01 & +0.18 [+0.13, +0.22] & 0.61 [0.58, 0.64] & +0.16 [+0.12, +0.21] \\
Slow motion & $r = 4$ & 0.84 $\pm$ 0.05 & 0.59--0.91 & 0.59 $\pm$ 0.02 & +0.25 [+0.21, +0.30] & 0.63 [0.60, 0.67] & +0.21 [+0.15, +0.25] \\
Slow motion & $r = 5$ & 0.87 $\pm$ 0.06 & 0.60--0.95 & 0.63 $\pm$ 0.01 & +0.25 [+0.20, +0.29] & 0.68 [0.64, 0.73] & +0.20 [+0.15, +0.24] \\
Slow motion & $r = 6$ & 0.90 $\pm$ 0.05 & 0.64--0.98 & 0.63 $\pm$ 0.01 & +0.27 [+0.24, +0.31] & 0.75 [0.70, 0.80] & +0.16 [+0.11, +0.20] \\
Slow motion & $r = 7$ & 0.92 $\pm$ 0.05 & 0.67--0.99 & 0.70 $\pm$ 0.01 & +0.23 [+0.19, +0.27] & 0.84 [0.80, 0.89] & +0.08 [+0.04, +0.13] \\
Slow motion & $r = 11$ & 0.93 $\pm$ 0.03 & 0.82--0.99 & 0.88 $\pm$ 0.01 & +0.05 [+0.02, +0.08] & 0.98 [0.96, 0.99] & $-$0.05 [$-$0.08, $-$0.03] \\
Slow motion & $r = 14$ & 1.00 $\pm$ 0.01 & 0.96--1.00 & 1.00 $\pm$ 0.00 & $-$0.004 [$-$0.006, $-$0.002] & 1.00 [1.00, 1.00] & $-$0.004 [$-$0.007, $-$0.002] \\
\bottomrule
\end{tabular}}
\end{table}

\begin{table}[h]
\centering
\caption{The coarse probes (window AUC, probe positive, 198 held-out windows; columns as in \tabref{tab:probes_fine}). Slow motion by speed: $\mathrm{idx}(t) = 4 + \lfloor v t \rfloor$; bounce: forward to the turning point $m$, then backward; shuffle: random permutation within consecutive blocks of $b$ generated frames. The null strengths ($v = 1$, $m = 14$, $b = 1$) are reality itself and are omitted. $v = 0.5$ and $v = 0$ are the same probes as $r = 7$ and $r = 14$ in \tabref{tab:probes_fine}.}
\label{tab:probes_coarse}
\small
\resizebox{\textwidth}{!}{%
\begin{tabular}{llcccccc}
\toprule
Probe & Strength & Trained (50 models) & Range & Untrained (5 seeds) & Trained $-$ untrained & Luminance & Trained $-$ luminance \\
\midrule
Slow motion & $v = 0.75$ & 0.81 $\pm$ 0.05 & 0.61--0.88 & 0.67 $\pm$ 0.02 & +0.14 [+0.09, +0.18] & 0.61 [0.58, 0.65] & +0.19 [+0.15, +0.24] \\
Slow motion & $v = 0.5$  & 0.92 $\pm$ 0.05 & 0.67--0.99 & 0.70 $\pm$ 0.01 & +0.23 [+0.19, +0.27] & 0.84 [0.80, 0.89] & +0.08 [+0.04, +0.13] \\
Slow motion & $v = 0.25$ & 0.93 $\pm$ 0.03 & 0.78--0.99 & 0.85 $\pm$ 0.01 & +0.08 [+0.05, +0.12] & 0.91 [0.87, 0.95] & +0.03 [$-$0.01, +0.06] \\
Slow motion & $v = 0$    & 1.00 $\pm$ 0.01 & 0.96--1.00 & 1.00 $\pm$ 0.00 & $-$0.004 [$-$0.006, $-$0.002] & 1.00 [1.00, 1.00] & $-$0.004 [$-$0.007, $-$0.002] \\
\midrule
Bounce & $m = 10$ & 0.74 $\pm$ 0.04 & 0.65--0.83 & 0.76 $\pm$ 0.01 & $-$0.02 [$-$0.05, +0.02] & 0.66 [0.63, 0.70] & +0.08 [+0.04, +0.11] \\
Bounce & $m = 7$  & 0.83 $\pm$ 0.04 & 0.74--0.92 & 0.82 $\pm$ 0.01 & +0.01 [$-$0.01, +0.04] & 0.80 [0.76, 0.85] & +0.04 [$-$0.01, +0.08] \\
Bounce & $m = 5$  & 0.81 $\pm$ 0.05 & 0.67--0.91 & 0.87 $\pm$ 0.01 & $-$0.05 [$-$0.08, $-$0.03] & 0.65 [0.62, 0.69] & +0.16 [+0.11, +0.21] \\
\midrule
Shuffle & $b = 2$  & 0.91 $\pm$ 0.04 & 0.68--0.96 & 0.57 $\pm$ 0.01 & +0.34 [+0.31, +0.37] & 0.69 [0.67, 0.73] & +0.21 [+0.18, +0.24] \\
Shuffle & $b = 3$  & 0.95 $\pm$ 0.03 & 0.78--0.98 & 0.75 $\pm$ 0.00 & +0.21 [+0.17, +0.24] & 0.80 [0.77, 0.84] & +0.15 [+0.11, +0.19] \\
Shuffle & $b = 5$  & 0.96 $\pm$ 0.02 & 0.88--0.98 & 0.80 $\pm$ 0.01 & +0.16 [+0.13, +0.20] & 0.89 [0.86, 0.93] & +0.07 [+0.03, +0.10] \\
Shuffle & $b = 14$ & 0.99 $\pm$ 0.00 & 0.98--1.00 & 0.97 $\pm$ 0.00 & +0.02 [+0.01, +0.03] & 1.00 [1.00, 1.00] & $-$0.004 [$-$0.009, $-$0.000] \\
\bottomrule
\end{tabular}}
\end{table}

\paragraph{Slow motion by speed.} The trained observer exceeds the untrained network at $v = 0.75$, 0.5 and 0.25 and the luminance control at $v = 0.75$ and 0.5; at $v = 0.25$ the interval of the difference to the luminance control includes zero ($+0.03$ [$-0.01$, $+0.06$]), and the freeze $v = 0$ is at ceiling for every observer. The coarse grid has no point between reality and $v = 0.75$ (0.81), which is what the fine grid supplies. The point $v = 0.75$ has four repeats, as $r = 4$ has, but at other positions ($t = 1, 5, 9, 13$ against $t = 1, 4, 8, 11$), and the AUCs differ (trained 0.81 against 0.84, untrained 0.67 against 0.59, luminance 0.61 against 0.63). The number of repeats is therefore one choice of strength axis among several, and the thresholds of \tabref{tab:thresholds} refer to the rule $\lfloor (14 - r)\, t / 14 \rfloor$.

\paragraph{Time reversal.} The bounce is the one family in which training adds nothing. At $m = 7$ and $m = 10$ the interval of the trained-minus-untrained difference includes zero, and at $m = 5$, the earliest turning point, the untrained network separates the probe better than the trained observer (0.87 against 0.81, difference $-0.05$ [$-0.08$, $-0.03$]). The trained curve is not monotone in $m$ (0.74, 0.83 and 0.81 at $m = 10$, 7 and 5) and never reaches 0.9. A likely reason is a property of the feature set. Each of the four statistics, the mean, the standard deviation, the lag-1 autocorrelation (computed about a common mean, with a numerator that is symmetric in time) and the mean absolute step-to-step change, is unchanged when a series is reversed in time. The response to a reversed stretch of stimulus is not the reversed response, but to the extent that it resembles the forward response played backwards the features cannot tell the two apart, and the bounce then differs from reality in them mainly at its turning point and through the frames it shows, repeating those before the turning point and omitting those after it (at $m = 5$ the continuation runs back into the context frames).

\paragraph{Shuffles within blocks.} Swapping within blocks of two already gives 0.91 on the trained responses, against 0.57 for the untrained network and 0.69 for the luminance control, and with no grid point between reality and $b = 2$ the coarse curve looks like a step; the pair swaps resolve it. With $b = 2$ each of the seven pairs is swapped with probability one half (a draw of the identity is redrawn), 3.6 pairs on average (3.61 $\pm$ 1.22 over the 198 windows), and its AUC lies between those of pair swaps at $k = 3$ and $k = 4$ for the trained observer (0.91 against 0.90 and 0.93) and for the luminance control (0.69 against 0.67 and 0.72), and slightly below that at $k = 3$ for the untrained network (0.57 against 0.58). For the trained observer it is slightly below the linear interpolation at 3.6 pairs (0.91 against 0.92), as expected for a mixture over $k$ of a curve that is concave in $k$. Larger blocks raise the AUC of all three observers, and the full permutation ($b = 14$) is at ceiling for all of them (0.97--1.00).

\subsection{World-model training and dream statistics}\label{app:training}

Across the five generator folds the training sets hold 3,309--3,585 frames (75 strip series of 25 sequences); the MDN-RNN's best validation epoch was 26, 50, 19, 28 and 36 and the RSSM's 58, 56, 55, 56 and 47 of 60. The persistence walk's $\sigma$ was 0.048, 0.062, 0.056, 0.061 and 0.057. Over the pooled held-out windows, the spatial standard deviation per frame is 0.164 for real frames, 0.170--0.208 for the persistence walk, 0.170--0.192 for the MDN-RNN and 0.127--0.150 for the RSSM (increasing with $\tau$ in every case); the persistence error, holding the last context frame, is 0.108.

\end{document}